\documentclass[10pt,journal,compsoc]{IEEEtran}
\usepackage{mathtools}
\usepackage{amsfonts}
\usepackage{amsmath}
\DeclarePairedDelimiter\floor{\lfloor}{\rfloor}
\newcommand\w{\boldsymbol{w}}

\usepackage{xcolor}
\usepackage{bbm}
\DeclareMathOperator*{\minimize}{Minimize}
\usepackage{nomencl}
\makenomenclature
\newcommand {\mX}{\boldsymbol{X}}
\newcommand {\mY}{\boldsymbol{Y}}
\newcommand {\mU}{\boldsymbol{U}}
\newcommand {\mE}{\boldsymbol{E}}
\newcommand {\mO}{\boldsymbol{O}}
\newcommand {\mP}{\boldsymbol{P}}
\newcommand {\mR}{\boldsymbol{R}}

\newcommand {\mQ}{\boldsymbol{Q}}
\usepackage{algorithm,algpseudocode}
\usepackage{array}
\newcolumntype{L}{>{\arraybackslash}m{0.55cm}}

\begin{document}

\bstctlcite{IEEEexample:BSTcontrol}

\title{Optimal User-Edge Assignment in Hierarchical Federated Learning based on Statistical Properties and Network Topology Constraints}

\author{Naram~Mhaisen, 
        Alaa~Awad~Abdellatif,~\IEEEmembership{Member,~IEEE,}
        Amr~Mohamed,~\IEEEmembership{Senior Member,~IEEE,}
        Aiman~Erbad,~\IEEEmembership{Senior Member,~IEEE,}
        and~Mohsen~Guizani, ~\IEEEmembership{Fellow,~IEEE,}% <-this % stops a space
\IEEEcompsocitemizethanks{\IEEEcompsocthanksitem N. Mhaisen, A.A. Abdellatif, A. Mohamed, and M. Guizani are with the Department of Computer Science and Engineering, College of Engineering, Qatar University, Qatar.\protect\\
E-mails: \{naram, aawad, amrm\} @qu.edu.qa; mguizani@ieee.org
\IEEEcompsocthanksitem A. Erbad is with the College of Science and Engineering, Hamad Bin Khalifa University, Qatar.\protect\\ E-mail: aerbad@ieee.org}% <-this % stops a space
}
\maketitle

\begin{abstract}
Distributed learning algorithms aim to leverage distributed and diverse data stored at users' devices to learn a global phenomena by performing training amongst participating devices and periodically aggregating their local models' parameters into a global model. Federated learning is a promising paradigm that allows for extending local training among the participant devices before aggregating the parameters, offering better communication efficiency. However, in the cases where the participants' data are strongly skewed (i.e., non-IID), the local models can overfit local data, leading to low performing global model. In this paper, we first show that a major cause of the performance drop is the weighted distance between the distribution over classes on users' devices and the global distribution. Then, to face this challenge, we leverage the edge computing paradigm to design a hierarchical learning system that performs Federated Gradient Descent on the user-edge layer and Federated Averaging on the edge-cloud layer. In this hierarchical architecture, we formalize and optimize this user-edge assignment problem such that edge-level data distributions turn to be similar (i.e., close to IID), which enhances the Federated Averaging performance. Our experiments on multiple real-world datasets show that the proposed optimized assignment is tractable and leads to faster convergence of models towards a better accuracy value.
\end{abstract}
% \begin{abstract}
% Learning-based applications have demonstrated practical use cases in ubiquitous environments and amplified interest in exploiting the data stored on users’ mobile devices. Distributed optimization algorithms aim to leverage such distributed and diverse data to learn a global phenomena by performing training amongst participating devices and repeatedly aggregating their local models’ parameters into a global model. Federated Averaging is a promising solution that allows for extending local training before aggregating the parameters, offering better communication efficiency. However, in the cases where the participants’ data are strongly skewed (i.e., local distributions are different), the model accuracy can significantly drop. To face this challenge, we leverage the edge computing paradigm to design a hierarchical learning system that performs Federated Gradient Descent on the user-edge layer and Federated Averaging on the edge-cloud layer. In this hierarchical architecture, the users might be assigned to different edges, leading to different edge-level data distributions. We formalize and optimize this user-edge assignment problem to minimize classes’ distribution distance between edge nodes, which enhances the Federated Averaging performance. Our experiments on multiple real datasets show that the proposed optimized assignment is tractable and leads to faster convergence of models towards a better accuracy value.
% \end{abstract}

\IEEEpeerreviewmaketitle

\newtheorem{theorem}{Theorem}
\newtheorem{lemma}[theorem]{Lemma}
\newtheorem{proposition}[theorem]{Proposition}

\section{Introduction}
% FL Motivation
Despite the increasing trend towards learning-assisted applications, the realization of these applications on end-user devices and data still poses significant challenges. Such challenges include the communication overhead required to move data to learning servers, in addition to privacy concerns associated with such movement \cite{kairouz_advances_2019}. Motivated by these challenges, local data storage and processing with global coordination is envisioned to be a promising approach in the future of computing \cite{8770530}. \emph{Federated Learning} (FL) is a manifestation of this approach that aims to enable end devices to learn highly-performing models without outsourcing their data or requiring high computational power \cite{mcmahan2017communication}. In FL, end devices execute parts of the learning algorithms on their local data, reaching a local model. The local models are then sent and synchronized (aggregated) in a cloud server, before being sent back to local devices. This process repeats until models converge, and contains multiple parameters, such as the number of local steps and the synchronization algorithm, which are usually tuned to maintain data privacy, avoid communication overhead, and reach overall higher accuracy models \cite{li2020federated}.

% EC => HFL
The paradigm of processing massive data generated by end devices is formalized in the networking community through the \emph {Edge Computing} (EC) architectures, where edge nodes such as home gateways, small-cells, or micro servers are equipped with storage and computation capabilities. Edge nodes communicate efficiently with end-user equipments (UEs), and they work with the remote cloud to perform large-scale distributed tasks that involve both local processing and remote coordination/execution \cite{8016573}. While EC is a general-purpose architecture that enables a form of hierarchical multi-layer processing, it can be used for learning tasks, leading to the Hierarchical Federated Learning (HFL) paradigm \cite{Client-Edge-Cloud}. In HFL, each UE performs local learning tasks leveraging its own dataset. The learned models are then periodically synchronized through an edge node, reaching an edge-level model. In turn, the edge nodes also periodically synchronize their models with the cloud server to reach the global model (i.e., cloud-level model).

%Non-iid challenge
In FL and HFL settings, the most widely used model synchronization algorithm is FedAvg \cite{mcmahan2017communication}. FedAvg uses stochastic gradient descent (SGD) as the learning algorithm. Hence, each UE performs several steps of SGD on its own data locally. Then, the local models are synchronized through calculating the average of model parameters, weighted by the size of their datasets. FedAvg proved effective in multiple learning scenarios, and in fact, is already used in production \cite{kairouz_advances_2019}. However, its performance degrades severely when the local datasets are highly imbalanced (i.e., skewed samples in each group of UEs) \cite{hsu2019measuring}. This is because gradient-based algorithms operate on the main assumption that data is independent and identically distributed (widely known as the i.i.d assumption). This assumption is violated in FL settings since each end device usually contains/generates skewed data depending on its task, specification, and characteristics \cite{kairouz_advances_2019}. Formally, the \emph{distribution over classes} differs significantly from a UE to another. Thus, each local dataset is often not a good representative of the global population and is a highly biased estimator of that population. In HFL, this challenge transfers to two levels: the distribution difference between UEs' datasets allocated to an edge node, and the distribution difference between the edge nodes' themselves \cite{Client-Edge-Cloud,wang2020towards}. 

There is limited room for improvement in solving the imbalance issue in the regular FL settings due to the inability to modify or share (parts of) users' data among themselves in order to reduce the distribution difference between users. However, such restrictions can be worked-around in the HFL settings; In many cases, UEs have access to more than one edge server; this is mainly due to network densification and the increased mobility of user devices. Such flexibility provides us with a vital control knob that we can leverage, which is deciding UE-edge assignments in such a way that minimizes the distribution difference between edge nodes (i.e., balances data under each edge). Such a network control problem is inline with the Reconfigurable Wireless Networks (RWNs) paradigm, which enables networks software, hardware, or protocols to be reconfigured to maximize a specific utility \cite{el-mougy_reconfigurable_2015}. For example, the optimization of edge-user assignments can be realized through reconfiguring routes for nodes' traffic to go through specific edges, ensuring balanced edge-level datasets for faster and more accurate learning. The formulation, and solution attempts, of edge-user assignment problem for data balancing are of utmost importance for the era of edge computing and learning-assisted applications. 

In this paper, we investigate the Hierarchical Federated Learning (HFL) architecture to get insights about its performance bounds and, more importantly, identify controllable parameters that affect the learning performance (i.e., learning speed and accuracy). As these parameters are shown to be mainly due to user-edge assignment, we optimize this assignment to enhance the distributed learning performance of UEs.

The main contributions are summarized as follows:
\begin{itemize}

    \item Analyzing the deviation of the learning parameters in the case of HFL with non-IID data from the parameters in the centralized benchmark case.

    \item Formalizing the problem of user assignment to edge nodes in HFL guided by the theoretical insights from the presented analysis.

    \item Proposing tractable solutions through the linearization of the problem with provable optimality in specific instances, and performant heuristics in the general case.
    
    \item Developing simulations based on real-world datasets and baseline neural architectures to demonstrate the effectiveness of the proposed optimized assignment approaches.
    
\end{itemize}

% \begin{itemize}
%     \item Formalizing the problem of user assignment to edge nodes in the Hierarchical Federated Learning settings guided by theoretical insights from Distributed Gradient Descent (DGD) algorithms.

%     \item Proposing tractable solutions through the linearization of the problem with provable optimality in specific instances, and performant heuristics in the general case.
    
%     \item Developing simulations based on real-world datasets and baseline neural architectures to demonstrate the effectiveness of the proposed optimized assignment approaches.

% \end{itemize}

The rest of this paper is organized as follows: Section \ref{sec:rw} explores related studies and positions the work among the FL optimization for non-i.i.d data literature. Section \ref{sec:sm} describes the general system model, and provide necessary theoretical background/insights for the problem formulation, which is introduced in section \ref{sec:pf}. Section \ref{sec:sol} proposes our solutions for this problem, which are empirically evaluated and discussed in \ref{sec:eval}, before concluding in Section \ref{sec:conc}.

\section{Related Work}
\label{sec:rw}
%Multiple directions
The Federated Learning paradigm includes multiple sub-problems that are being actively investigated in the literature. For instance, Authors in \cite{wang2019adaptive} focused on optimizing the global synchronization (aggregation) frequency based on the resources of the learning devices. Secure synchronization techniques are explored in \cite{45808}, where cryptographic multi-party computation is used to limit the server's ability to infer users' data from the models. The authors in \cite{Nishio2019ClientSF} proposed methods to select only a subset of users for the synchronization process and demonstrated the advantages of such techniques on the learning speed and performance, especially in the presence of resource-constrained nodes. The authors in \cite{tran_federated_2019} formulated a multi-objective optimization of minimizing energy cost and learning time for wireless devices given a target model accuracy, and possibly heterogeneous computational and communication capabilities. However, the aforementioned studies do not directly propose methods to alleviate the effects of non-i.i.d data. Thus they are orthogonal to our work in this paper.

%% Our direction - non-i.i.d
Some works identified and primarily addressed the non-i.i.d issue in FL. For example, Zhao et al. \cite{zhao_federated_2018} determined, theoretically and empirically, the effect of local data skewness on the global learning model. Since exchanging data between users is not possible, the authors proposed to use a central proxy iid dataset, which is distributed over clients. Mixing this data with the local user's data will make it less biased and potentially a better estimator of the overall dataset. Although the experiments demonstrated the effectiveness of this proxy dataset, obtaining such a dataset is not always feasible. Furthermore, the communication overhead of downloading parts of this data to each user is a considerable disadvantage. This line of work was improved upon by \cite{wang_optimizing_2020}, where the authors avoided needing proxy data through carefully selecting users participating in each synchronization round. The selection process is done through inferring the distribution over labels at each user's dataset, and then selecting a group of clients that are as diverse as possible, which results in a better learning performance due to the better quality data samples. Alternatively, \cite{sattler_robust_2019} focused on the aggregation technique itself. The authors proposed to replace the averaging with a compression-based aggregation, which alleviates the weights divergence effects that would otherwise be more pronounced in the averaging. These studies have shown the effect of data distribution on the obtained performance. However, they neither consider the communication overhead nor the HFL architecture and its associated user-edge assignment problem.   

% More specific: non-i.i.d & HFL, jointly
There are few studies that jointly consider HFL for non-iid data. A notable example is \cite{Client-Edge-Cloud}. In this paper, the authors extended the results of \cite{wang2019adaptive} and generalized it to HFL. They analytically linked learning performances (as modeled by the gap between the federated model and centralized virtual model) with both: the client-edge distribution difference and the edge-cloud distribution difference. This line of work was further improved in \cite{wu_accelerating_2020} with enhanced probabilistic client selection that is set to avoid failed/slow devices and their potential impact on the learning process (e.g., halting). While Important insights about user selection/grouping were obtained from the theoretical analysis of these studies, an optimization formalism was not provided. In contrast, the client selection proposed in \cite{9145182} considered the distribution distance as a part of an optimization problem. However, the distance is optimized through adjusting the local batch size, which might lead to the underutilization of data in strongly skewed users. In \cite{duan_self-balancing_2021}. A hierarchical architecture that includes mediators, which can be considered as edge nodes, is proposed. Also, an assignment scheme that aims to minimize the KL divergence (a distribution distance measure) of edges distributions is utilized. Our work differs in that the used measure for distribution distance is the absolute value of the distribution difference (also known as Earth Moving Distance) since it is the one that shows in the theoretical analysis. The authors did explicitly consider the mediator-client assignment constraints as mediators were assumed to always be available. Lastly, sequential gradient descent was proposed in the user-mediator layer, which necessitates a potentially expensive data augmentation to balance individual user's contributions to the global model  \cite{aug-1811-11479}. To the best of our knowledge, formulating and optimizing UEs' assignment to edge servers for hierarchical federated learning based on both statistical properties of local datasets, and communication constraints have not been addressed in the literature.
%% Note: a possible critic of user selection is ignoring biased users in duan

%% FedSGD in Datacenter
Different from distributed optimization on mobile devices, distributed optimization is also done on servers at data centers. It was demonstrated in \cite{wang2019adaptive} and \cite{45187} that synchronous distributed gradient descent algorithms for distributed optimization (also referred to as FedSGD \cite{mcmahan2017communication}) delivers identical performance to an SGD optimization done on a virtually aggregated local datasets. FedSGD is a gradient-based algorithm equivalent to federated learning when the global aggregation occurs at every step, and every local step is a full deterministic gradient. FedSGD is robust to non-iid data distribution. However, this comes at the cost of more communications rounds since aggregation is done after each local gradient step. Thus, more communication rounds are needed. Nonetheless, as indicated in \cite{lim_federated_2020}, synchronous distributed gradient descent can be exploited whenever the communication links allow (e.g., in data centers).

In this paper, we design an optimized hierarchical learning system considering both the statistical properties of local datasets and wireless communication connectivity. Our work focuses on the scenario where edge nodes are close to UEs and thus have low communication cost and delay. Hence, we adopt FedSGD in the UE-Edge layer. On the other hand, edge-cloud communication is delay-prone. Therefore, we adopt FedAvg, which performs a higher number of local steps before the cloud aggregation. In this context, we optimize the UE-edge assignment in order to minimize the communication rounds between edge nodes and the cloud. The overall objective of our optimization problem is to achieve high accuracy with the minimum delay as majorly influenced by the edge cloud communication rounds.

\section{System Model}
\label{sec:sm}

\subsection{HFL architecture}
\begin{figure}[t]
\centering
\includegraphics[width=0.485\textwidth]{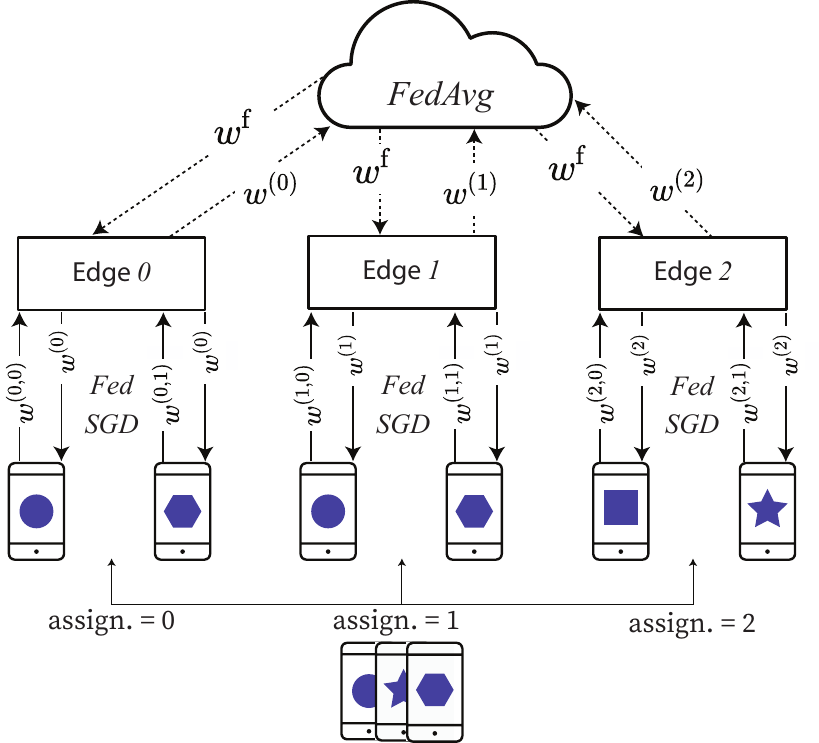}
\caption{The hierarchical learning architecture. UEs are assigned to an edge node for learning the edge-specific model. Edge nodes collectively learn the global model with the help of a cloud server. The shapes on the UEs represent the class of that UE's data (assuming an extreme case of one class per UE).}
\label{fig:Sys}
\end{figure}

We consider a hierarchical User-Edge-Cloud learning system architecture shown in Fig. \ref{fig:Sys}. The UE data is represented by the matrix $\boldsymbol{U}^{(C\times U)}$ where $C$ is the number of classes in the learning problem, $U$ is the total number of UEs, and an entry $\mU_{c,u}\in\mathbb{N}$ is the number of instances of class $c$ in the data of user $u$. The Network Topology Constraints (NTCs) refer to the connectivity constraints between the UEs and the edge nodes. NTCs are represented by $\boldsymbol{R}^{(N\times U)}$, where the number of edge nodes is $N$, and an entry $\mR_{n,u}\in\{0,-1\}$ represents whether edge $n$ is in the communication range of user $u$ (entry value of $0$), or $n$ is not in reach of $u$ (entry value of $-1$). 

Upon a specific assignment, we refer to the dataset of each UE by $\mathcal{D}^{(n,u)}$, which is the $u$-th column of $\mU$. The virtual dataset of an edge $n$ is $\mathcal{D}^{(n)} ={\cup_{u}\mathcal{D}^{(n,u)}}$, and the virtual global aggregated dataset as $\mathcal{D}={\cup_{n,u}\mathcal{D}^{(n,u)}}$. Note that we use ``virtual'' since these datasets do not get physically stored on the edge or the cloud. However, we define them for analysis. Each UE group performs FedSGD learning with its edge node. Under this learning scheme, each UE maintains a set of learning parameters (UE weights) $\w^{(n,u)}$ that are optimized based on the dataset $\mathcal{D}^{(n,u)}$. The UE weights are aggregated at every communication round to form the edge node parameters $\w^{(n)}$. Then, edge nodes synchronize their parameters at the cloud through the FedAvg learning scheme (i.e., aggregation at a pre-determined frequency). The process repeats until convergence.

\subsection{Mathematical demonstration}
In this section, we present the formulas that define the learning  process. We use the widely adopted cross-entropy loss function:  
\begin{align}
\label{eq:cross_ent_loss}
&\mathcal{L}(\w) &=& \mathbb{E}_{y\sim p(y),\boldsymbol{x}\sim q(\boldsymbol{x}|y)} \left[ -\mathbbm{1}_{ y=i} \log d_i(\boldsymbol{x};\w) \right] \nonumber \\
&&=&\sum_{i=1}^C -p(i) \mathbb{E}_{\boldsymbol{x}\sim q(\boldsymbol{x}|y=i)} \left[ \log d_i(\boldsymbol{x};\w) \right]
\end{align}
Where $\boldsymbol{x}$ is a feature vector of a data point, $p(\cdot)$ is the global classes distribution (global distribution), $q(\cdot|\cdot)$ is the likelihood function, and $d_i(\boldsymbol{x},\w)$ is the probability of the $i$-th class for input $\boldsymbol{x}$ under parameters $\w$.
The objective of the overall learning process is to reach a set of parameters $\w$ that minimizes the loss across virtual global dataset:

\begin{equation}
\label{eq:learn_objective}
\displaystyle{\minimize_{\w} \sum_{i=1}^C -p(i) \mathbb{E}_{\boldsymbol{x}\sim q(\boldsymbol{x}|y=i)} \left[ \log d_i(\boldsymbol{x};\w) \right]}
\end{equation}
The loss function in (\ref{eq:cross_ent_loss}) can be minimized using the regular gradient descent update:
\begin{equation}
\label{eq:centr_grad_update}
\w_t = \w_{t-1}  -\delta \nabla_{\w}\mathcal{L}(\w)
\end{equation}
We refer to the parameters in (\ref{eq:centr_grad_update}) as the centralized model. Since the data is distributed across different end-users, the loss function in (\ref{eq:centr_grad_update}) cannot be evaluated directly. Instead, each end-user device $u$ locally computes the loss function over its own data samples, forming a set of local parameters $\w^{(n,u)}$, which are optimized through the gradients steps: 

\begin{equation}
\w^{(n,u)}_t = \w^{(n,u)}_{t-1}  -\delta \nabla_{\w^{(n,u)}}\mathcal{L}^{(u)}(\w^{(n,u)}) 
\end{equation}

\begin{align}
&\mathcal{L}^{(u)}(\w) &=& \mathbb{E}_{y\sim p^{(u)}(y), \boldsymbol{x}\sim q(\boldsymbol{x}|y)} \left[ -\mathbbm{1}_{ y=i} \log d_i(\boldsymbol{x};\w) \right] \nonumber \\
&&=&\sum_{i=1}^C -p^{(u)}(i) \mathbb{E}_{\boldsymbol{x}\sim q(\boldsymbol{x}|y=i)} \left[ \log d_i(\boldsymbol{x};\w) \right]
\end{align}
Where $p^{(u)}(\cdot)$ is the user-specific classes distribution, which is different across users due to the different classes proportions and different dataset sizes as well. In the hierarchical learning settings, the synchronization (i.e., aggregation) of learners' weights is periodically done by taking the average of the parameters weighted by the local dataset's size. The learners use the averaged parameters until the next aggregation round. Specifically, in HFL, the weights are synchronized across all users belonging to a specific edge every $T'$ local gradients steps, and then synchronized across all edges every ($T'\times T$) steps. Hence The parameters of an edge node at the $m$-th \emph{edge aggregation} are:

\begin{equation}
\w^{(n)}_{m\times T' } =\sum_{u=1}^U r^{(n,u)}\w^{(n,u)}_{m\times T' }
\end{equation}
where
\begin{equation}
    r^{(n,u)} =\frac{|\mathcal{D}^{(n,u)}|}{ |\cup_u\mathcal{D}^{(n,u)}|}
\end{equation} 
Similarly, at each $m-th$ \emph{cloud aggregation}, with aggregation frequency $T$, the weights are averaged across all users in all edges:
\begin{equation}
\w^{\text{f}}_{m(T'  \times T)} =\sum_{n=1}^Nr^{(n)}\w^{(n)}_{m(T'\times T)}
\end{equation}

where:
\begin{equation}
    r^{(n)} =\frac{|\mathcal{D}^{(n)}|}{ |\cup_n\mathcal{D}^{(n)}|}
\end{equation} 

$\w^{\text{f}}_{m(T'  \times T)}$ are the parameters averaged across all edges. These are referred to as the federated weights, and they represent the final model that is learned and used by UEs.

Generally, we are interested in the term shown in (\ref{eq:divergence}), which represents the divergence between the federated weights and the central weights. Note that the central weights represent the parameters reached in the hypothetical case wherein the datasets of all UEs are aggregated at a central node, and the conventional centralized gradient descent is used. We define these weights for analysis only (i.e., in order to quantify the deviation between FL's weights and central weights). 
\begin {equation}
\label{eq:divergence}
\|\w^{\text{f}}_{m(T \times T')} - \w_{m(T \times T')}\|
\end{equation}
The divergence expression in (\ref{eq:divergence}) represents the deviation caused due to the distribution of the datasets and performing learning locally. Lower values are desired since they represent a closer model to the centralized one.

%% %TODO Should i add a workflow section here with an algorithm here, probably ? 

\section{Problem Formulation}
\label{sec:pf}
In order to minimize the divergence in (\ref{eq:divergence}), we need an expression that links it with controllable knobs (i.e., decisions variables that we can optimize). To that end, we first present theoretical lemmas that help us reach the target expression and provides us with insight on factors that affect this divergence. Then, we optimize these factors with the aim of minimizing the divergence.
% In order to minimize the divergence in (\ref{eq:divergence}), an expression that links it to controllable decision variables is needed. To that end, we present theoretical lemmas that help us reach the target expression and provides us with insight on factors that affect this divergence. 

%=\cup_u\mathcal{D}^{u\in\mathcal{U}_n}
\begin{lemma}
Given an edge node $n$ and a set of learning UEs assigned to it $\mathcal{U}_n$. Then, if the synchronization period $T'$ is set to one, the federated weights reached after $t$-th aggregation, $\w^{(n)}_{m}$, are equal to the weights reached by using centralized gradient descent on the virtual aggregated dataset. $\mathcal{D}^{(n)}$. 
\begin {equation}
\sum_{u\in\mathcal{U}_n} r^{(n,u)}\w^{(n,u)}_{t } = \w^{(n)}_{t-1} - \delta    \nabla \mathcal{L}^{(n)}(\w^{(n)}_{t-1}), \forall t, n
\end {equation}
\end{lemma}

\begin{IEEEproof}
The expression on the left-hand side represents the FedSGD aggregation rule. The lemma holds due to the linearity of the gradient operator. The details are provided in Appendix A.
\end{IEEEproof}

Lemma-1 states that when we perform edge aggregation after every local gradient step, the distributed SGD is equivalent to the centralized gradient descent, where the latter assumes that all data samples are available at a centralized location and the global loss function and its gradient can be calculated directly. The role of this lemma is to demonstrate that in the hierarchical architecture illustrated earlier in Fig. \ref{fig:Sys} there will be no user-edge parameter divergence, and the parameters divergence will be entirely due to the averaging of $\w^{(n)}$ (i.e., edge-cloud divergence).
Setting $T'=1$ is possible when the distance between the edge nodes and mobile devices is small, and the expected delay is limited, which is the case focused on in this paper. 

Unlike user-edge communication, edge-cloud communication is expensive in terms of delay. Thus, each edge node should perform several steps of local updates before communicating, aggregating only every $T$ steps. Such an aggregation period induces a divergence between the aggregated (federated) weights, and their centralized virtual counterparts. The next lemma provides an expression for such divergence.   

\begin{lemma}
Given $N$ edge nodes, each with a distribution $p^{(n)}$ over $C$ classes in a federated learning setting with a synchronization period $T$. Then, assuming that $J_i(\w)$ is $L_i$-Lipschitz, the following inequality represents an upper bound on the divergence between the federated weights, and the virtual central weights after the $m$-th synchronization:
\begin{multline}
\lVert\w^{\text{f}}_{mT}-\w_{mT}\rVert \leq
\sum_{v=0}^{m}  \left( \sum_{n=1}^N  r^{(n)} (P^{(n)})^{T} \right)^v \nonumber \\
\times \delta  \sum_{n=1}^N  r^{(n)} \sum_{h=0}^{T-2}   (P^{(n)})^{h+1}   J_{max}(\w_{(m-v)T-2-h}) \times \|D^{(n)}\|_1
\end{multline}
where
\begin{align}
&P^{(n)}=1 + \delta \sum_{i=1}^C p^{(n)}(i) \times L_i \label{eq:pn}\\ 
&D^{(n)} = \{ \lVert p(i) -  p^{(n)}(i) \lVert \}_{i=1}^C\label{eq:Dn}\\
&J_{i}(\w) = \nabla_{w} \mathbb{E}_{\boldsymbol{x}\sim q(\boldsymbol{x}|y=i)} \left[ \log d_i(\boldsymbol{x};\w) \right]\label{eq:Ji}\\
&J_{max}(\w) = \max_{i\in{1,\dots,C}} \lVert J_i(\w) \rVert\label{eq:Jmax}
\end{align}

\end{lemma}
(\ref{eq:pn}) is a positive constant that represents the dot product between each edge's classes' distribution vector and the classes' Lipchitz-constants vector, plus $1$. (\ref{eq:Dn}) is the distance vector for edge $n$, where the $i$-th element is the distance between the probability of the class $i$ per edge $n$'s distribution and the probability of that class per the global distribution, (\ref{eq:Ji}) is the negative gradient of the loss function, and (\ref{eq:Jmax}) is the greatest norm across all gradient vectors. This is used to bound all the terms that contain the gradient vector $J_i(\cdot)$, which repeatedly appear in the analysis.

\begin{IEEEproof}
The proof is provided in Appendix B. We follow stages similar to those in \cite{zhao_federated_2018} for regular FL, which is based on distributed SGD analysis. However, we adapt it to HFL settings and extend it to be written entirely in terms of $D^{(n)}$, which is necessary for our optimization formulation.

\end{IEEEproof}

Lemma-2 provides the eventual weight divergence induced due to the federation of the learning process (i.e., due to performing local steps on each edge node and aggregating the weights each $T$ steps). It also gives an upper bound, hereafter referred to as ($\beta$), in terms of variables that we can control; Namely, the proportion of each edge dataset to the global dataset $r^{(n)}$, and its class distribution distance from the global aggregated dataset $\|D^{(n)}\|_1$. Given that $\beta$ is a sum of positive terms, each of which is multiplied by $r^{(n)}$ and $\|D^{(n)}\|_1$, we have the remark that
\begin{equation}
    \label{eq:UB_prop}
    \beta \propto \sum_{n=1}^N  r^{(n)} \times \|D^{(n)}\|_1
\end{equation}

$r^{(n)}$ and $\|D^{(n)}\|_1$ are controllable knobs since we can allocate users to different edge nodes, which would result in different class distributions (hence different values of $\|D^{(n)}\|_1$) and different edge dataset sizes (hence different values of $r^{(n)}$) according to our assignment. Thus, we aim to find an assignment that minimizes the term in (\ref{eq:UB_prop}), to tighten the divergence of the federated weights from the virtual central weights.

In order to calculate $\|D^{(n)}\|_1$, the data distribution at users is needed. While this does reveal some information about the user, the core FL requirement of not moving users' data is still met. Nonetheless, the optimizing server may assume, approximate, or learn the distribution of users who opt-out of sharing this information. Overall, we are interested in minimizing the divergence bound given (an approximation of) data distributions.

\subsection{Optimization Formulation}

Using the notation in Table \ref{tab:notation}, we formulate an optimization problem whose decision variables are UE-edge assignments with the objective of minimizing the upper bound of weight divergence in (\ref{eq:UB_prop}). 

%(node since the left multiplication by the row vector of ones yields the sum of rows of the multiplied matrix)

\begin{table}
\renewcommand{\arraystretch}{1.2}
    \caption{Notation}
    \label{tab:notation}
    \centering
    \begin{tabular}{c|>{\centering\arraybackslash}m{6cm}}
            \hline
         Notation & Meaning   \\ \hline
         
         $D^{(n)}$ & A vector of classes' probability distances between edge $n$ and the global population \\
         $r^{(n)}$ & The proportion of edge $n$ dataset size to the global dataset size \\
         $C$ & Number of classes \\
         $N$ & Number of edge nodes \\
         $U$ & Number of end users \\
         $O$ & Number of user groups \\
         $\boldsymbol{E}^{(C\times N)}$ & Classes distributions per edge \\
         $\boldsymbol{U}^{(C\times U)}$ & Classes distributions per user \\
         $\boldsymbol{O}^{(C\times O)}$ & Classes distributions per group\\
         $\boldsymbol{P}^{(C\times 1)}$ & The global Classes distribution \\
         $\boldsymbol{X}^{(U\times N)}$ & User-edge assignments \\
         $\boldsymbol{Y}^{(O\times N)}$ & Group-edge assignments \\
         $\mR^{(N\times U)}$ & the edge-user communication range indicator matrix\\
         $\boldsymbol{Q}^{(N\times O)}$ & the edge-group communication range indicator matrix \\ \hline
    \end{tabular}

\end{table}

Since $\mX$ is a matrix of binary decision variables (whenever $\mX_{u,n}=1$, user $u$ is assigned to edge $n$). The multiplication $\mU\mX$ assigns each edge to a specific distribution of classes. In addition, $\boldsymbol{1} (\mU \mX)$ gives the total number of data points in every edge, since $\boldsymbol{1}$ is a row vector of ones which sums the rows of the multiplied matrix. $\mathbf{diag}^{-1}(\boldsymbol{1}(\mU \mX))$ is a matrix whose $n$-th diagonal element is the total number of data points in the $n$-th edge, where $diag^{-1}(\cdot)$ is an operator which takes a vector and returns a diagonal matrix whose diagonal is the reciprocal of the elements in that vector. We can then define the matrix  $\mE = \mU\mX\times\mathbf{diag}^{-1}(\boldsymbol{1}(\mU \mX))$, which represents the proportion of each class $c$ in each edge $n$ (i.e., $\mE_{c,n} = p^{(n)}(c)$). This is because the presented expression for $\mE$ divides the total number of each class's instances in edge $n$ on the total number of data points in that edge. Using these structures, we can write
\begin{align}
    \boldsymbol{D}=&\{\|D^{(n)}\|_1\}_{n=1}^{N} \nonumber \\ 
    =&  \left( \boldsymbol{1} \times | \mU \mX\times\mathbf{diag}^{-1}(\boldsymbol{1} (\mU \mX))-\mP_N | \right)
\end{align}
where $\boldsymbol{P}_N$ is obtained by replicating the column vector $\boldsymbol{P}$ into $N$ columns. Thus, our objective function can be written as:
\begin{align}
   \boldsymbol{\theta}=& \sum_{n=1}^N  r^{(n)} \times \|D^{(n)}\|_1\label{eq:wavg_dist}\\
    =&  (\boldsymbol{1} (\mU \mX)) \times \left( \boldsymbol{1} \times | \mU \mX\times\mathbf{diag}^{-1}(\boldsymbol{1} (\mU \mX))-\mP_N | \right)^{\top}\nonumber
\end{align}
Hence, our optimization problem can be formulated as:

\begin{align}
\boldsymbol{P_o}:  \minimize_{\mX}  \quad & \boldsymbol{\theta} \nonumber \\
\text{Subject to} \quad& \sum_n \mX_{u,n} =1, \forall u && \label{orig_single_allocation}\\
\quad& \mX \mR \geq 0 \label{orig_proximity}\\
\quad& \mX_{u,n} \in \{0,1\} \label{orig_binary_decisions}
\end{align}

The constraint in (\ref{orig_single_allocation}) ensures a user is allocated to a single edge node, and constraint (\ref{orig_proximity}) ensures that the users are allocated to the edges in their communication range.  (\ref{orig_binary_decisions}) is used to express a binary decision variable.
The above problem is a non-linear integer program \cite{belotti2013mixed}, which is known to be an NP-hard problem (the problem is akin to the Generalized Assignment Problem \cite{kellerer_knapsack_2004}). The non-linearity is because $\mE$ depends on $\mX$ in a non-linear way due to the decision variables $\mX$ appearing in the numerator and denominator of $\mE$ entries. To tackle this problem, we propose two solution approaches. The first solution is based on transforming the formulated problem into a linear integer program. While this transformation does not solve the NP-Hardness of the problem, it enables us to utilize several known approximations that are only feasible in the linear case. The second approach is an equalization heuristic, which is based on the observation that it provides an optimal solution in a particular case, as will be seen in the next section.

%%%%%%%%%%%%% Switch to configurations
\section{Proposed solutions}
\label{sec:sol}
This section presents the proposed two approaches for solving our optimization problem that aims to minimize the distribution difference between edge nodes and the global one given a set of connectivity constraints. The optimization is performed by the cloud server, and the resulting solution, i.e., the edge-user assignments, are communicated by the servers to the user devices so that they determine their edge node prior to starting the training. Note that it is possible that while the training is being executed, a change in the system occurs. For example, a new edge node joins, or a new NTC arises. In such a case, the optimization can be done again, depending on the change's expected effect. The recurrence of the optimization can be determined based on the complexity of the learning problem as it influences the learning duration.

\subsection{Problem transformation}

To tackle the NP-Hardness of the formulated problem, we leverage the following transformations and assumptions. First, users who have similar local datasets (i.e., class distribution) and in the proximity of the same edge nodes are grouped into one group of a specific size. This enables us to replace the user-state matrix $\mU$ by the group-state matrix $\mO$ whose entries $\mO_{c,o}$ are the number of class $c$ instances in group $o$. We define the vector $\boldsymbol{G}$ whose elements, $G_o$ are the number of users within group $o$ (the size of the group). Accordingly, $\mX$ is replaced by  $\mY, \mY_{o,n}\in\mathbb{N}$. Note that the dimension of $\mY$ is smaller than that of $\mX$ (as the group contains more than one user). Hence, this assumption reduces the search space of the decision variable. Also, the products $\mU \mX$ and $\mO \mY$ still have the same dimension. The NTCs matrix is also redefined to be per group $\boldsymbol{Q}^{(N\times O)}$.

The second step is to linearize the objective function through introducing the following constraint:
\begin{equation}
    \sum_c (\mO \mY)_{c,n} = S, \forall n
\end{equation}

%% used to follow from the assumption $\mathbf{diag}^{-1}(\boldsymbol{1}(\mO \mY) ) = \frac{1}{S}$.
which implies that the sizes of edge's datasets are equal. One case that satisfies this assumption is when the local dataset sizes are equal, and the number of users allocated to every edge is also equal. It follows from the assumption that:
\begin{equation}
\label{eq:simpl_1}
 \boldsymbol{1} \mO \mY  = [S, S, \dots, S]_{(1 \times N)} = \boldsymbol{S}
 \end{equation}
 \begin{equation}
 \label{eq:simpl_2}
 \mO \mY \times \mathbf{diag}^{-1}(\boldsymbol{1} \mO \mY)= \frac{1}{S} \mO\mY.
 \end{equation}
(Note that the decision variable $\mY$ is removed from the denominator of (\ref{eq:simpl_2})).
By replacing the product $\mU\mX$ by $\mO\mY$ and substituting (\ref{eq:simpl_1}) and (\ref{eq:simpl_2}) in the original optimization problem, we obtain the following objective function 
\begin{align}
    &\boldsymbol{\hat{\theta}} =
    \boldsymbol{S} \left( \boldsymbol{1} \times
        | \frac{1}{S}\mO \mY-\mP_N | \right)^\top
\end{align}
Thus, our optimization problem is transformed into:  
\begin{align}
\boldsymbol{P_1}: \minimize_{\mY}  \quad & \boldsymbol{\hat{\theta}} \label{eq:lad} \\
\text{Subject to} \quad& \sum_n \mY_{o,n} \leq G_o, \forall o &&\label{eq:configs_budget_const}\\
\quad& \sum_c (\mO \mY)_{c,n} = S, \forall n \label{eq:equal_dp_const} \\ 
% \quad& \sum_i\mY_{i,j}\times |\mathcal{D}_{u}| = S \label{eq:equal_dp_const}, \forall j\\ 
\quad& \mY\mQ \geq 0 \label{eq:prox_const}\\ 
\quad & \mY_{o,n} \in \mathbb{N}\label{eq:int_allocations_const}
\end{align}

Constraint (\ref{eq:configs_budget_const}) ensures that the users' assignment from each group does not exceed the group's size. Constraint (\ref{eq:equal_dp_const}) ensures the equal total number of data samples allocated to each edge, which is the assumption that enabled us to linearize the problem. Constraint (\ref{eq:prox_const}) is ensuring that users are allocated to edges in their communication ranges. Finally, constraint (\ref{eq:int_allocations_const}) represents the integer decision variables. Since $\boldsymbol{P_1}$ was obtained by introducing a linear constraint to $\boldsymbol{P_0}$, a feasible solution for $\boldsymbol{P_1}$ is also feasible for $\boldsymbol{P_0}$\footnote{Since the decision variable is changed from $\mX$ to $\mY$, this statement holds when a one-to-one mapping between them is provided (in our case, each user in $\mX$ can be mapped into an independent group in $\mY$ whose size is one)}.

For solving the problem $\boldsymbol{P_1}$, which is a Least Absolute Deviation (LAD) problem, we leverage the epigraph technique with a change of variables \cite{boyd2004convex}. Indeed, a slack decision variable $\boldsymbol{\mu}$ is defined to allow for transforming $\boldsymbol{P_1}$ into an equivalent linear program, as follows: 

Transformation into linear program:

\begin{align}
\boldsymbol{P_2: } &\minimize_{\boldsymbol{\mu},\mY} && \boldsymbol{\boldsymbol{S}\mu^\top}&&\label{eq:lp}\\
&\text{Subject to} & \boldsymbol{\mu} &\leq \boldsymbol{1}\left(\frac{1}{S}\mO \mY-\mP_N \right) \label{eq:const_abs1}&&\\
&&-\boldsymbol{\mu} &\leq \boldsymbol{1}\left(\frac{1}{S}\mO \mY-\mP_N \right) \label{eq:const_abs2}&&\\
&& (\ref{eq:configs_budget_const})&,  (\ref{eq:equal_dp_const}), (\ref{eq:int_allocations_const}), (\ref{eq:prox_const}) &&\nonumber
\end{align}

The introduced constraints in (\ref{eq:const_abs1}) and (\ref{eq:const_abs2}) are necessary for the transformation and come from the removal of the absolute value from the objective function. Although the problem $\boldsymbol{P_2}$ is still an integer program whose complexity is NP-hard, it can be approximately solved in an efficient way compared to the original problem $\boldsymbol{P_0}$. This is due to the program's linearity, which enables solution heuristics, such as relaxing the integrality and rounding the solutions (stochastic, or deterministic rounding), or solving the problem directly through Branch and Bound search but with improved bounding \cite{kochenderfer2019algorithms}. If the latter is used, then the bounding is done through leveraging the continuous version of the linear program, which can be solved easily, to calculate the lower bound of each node, enhancing the search process compared to the regular Branch and Bound algorithm. These linear programming techniques are readily available in most of the current solvers.

\subsection{Heuristic}

\begin{algorithm}
\caption{Equal assignment}
\label{alg:heu}
\begin{algorithmic}[1]
 \renewcommand{\algorithmicrequire}{\textbf{Input:}}
 \renewcommand{\algorithmicensure}{\textbf{Output:}}
\Require $\mO$ The groups' distribution matrix, $\boldsymbol{G}$ The groups' sizes. 
\Ensure $\mY$: The assignment decisions.

\State \textbf{for} each group $o$ \textbf{do}
\State \hskip1em $\boldsymbol{N} = \{n|Q_{n,o}=0\}$
\State \hskip1em $\rho=|\boldsymbol{N}|$

\State \hskip1em \textbf{if} $G_o\mod\rho=0$: 
\State \hskip2em $\mY_{o,n}=\frac{G_o}{\rho}, \forall n$
\State \hskip1em \textbf{else if} $G_o>\rho$: 
\State \hskip2em $\mY_{o,n}=\floor{G_o/\rho}, \forall n$
\State \hskip2em Add $(G_o\mod\rho)$ to $\mY_{o,|\boldsymbol{N}|}$
\State \hskip1em \textbf{else if} $G_o<\rho$: 
\State \hskip2em $\mY_{o,n}=1, \forall n \leq G_o$
\State \hskip2em $\mY_{o,n}=0, \forall n > G_o$

\end{algorithmic}
\end{algorithm}

The optimization formulation in $\boldsymbol{P_1}$ can provide us with optimal solutions for problems of small size, which is useful for performance comparison. However, to solve large real-world instances of $\boldsymbol{P_0}$, a more efficient method is desired to avoid the exponential solution space of integer programs. We design a solution algorithm based on the simple observation that, in some instances, the problem $\boldsymbol{P_0}$ can be solved optimally. Specifically, when the number of available groups of users can be equally divided among edge nodes in their communication range. In this context, we propose the greedy algorithm shown in Algorithm \ref{alg:heu}. It aims to distribute users' groups equally among the edges in their communication range. If an equal assignment is possible (line $4$), the optimal solution for $\boldsymbol{P_1}$ and $\boldsymbol{P_0}$ will be guaranteed (as will be shown in the evaluation). If a group's size is not divisible by the number of edges (line $6$), the group is divided equally among edge nodes, while the remainder is assigned to a remaining edge node. Lastly, in the case where the size of a user group is less than the number of edges, the available users can be assigned to any subset of the edge nodes. 

\begin{lemma} If $\rho$ divides $G_o$, then the following solution is optimal:
\begin {equation}
\mY^*_{o,n}=\frac{1}{\rho}G_o, \forall n
\end{equation}
\label{lemm:opt_alloc}
\end{lemma}
This lemma states that if the users' group sizes are divisible by the number of edges, and users can be assigned to any of the edge nodes, then the equal assignment is optimal.

\begin{IEEEproof}
\textit{Constraints satisfaction: }
$\mY^*$ satisfies constraint (\ref{eq:configs_budget_const}) since $\rho \geq 1$. It also satisfies constraint (\ref{eq:equal_dp_const}) since
\begin{align}
     S &= \sum_c (\mO \mY)_{c,n} \nonumber \\
     &= \sum_c \left(\sum_o\mO_{c,o}\times\frac{1}{\rho}G_o\right) \nonumber
\end{align}
which is the same for all nodes. Constraints (\ref{eq:prox_const}) and (\ref{eq:int_allocations_const}) are satisfied by the construction function.

\textit{Function value: }
\begin{align}
    \frac{1}{S}\mO\mY_{c,n}&=\frac{\sum_o\mO_{c,o}\times\frac{1}{\rho}G_o}{\sum_c \left(\sum_o\mO_{c,o}\times\frac{1}{\rho}G_o\right)} \nonumber \\
    &=\frac{\sum_o\mO_{c,o}\times G_o}{\sum_c (\sum_o\mO_{c,o}\times G_o)} \nonumber \\
    &= p(c)
\end{align}
 We can then write
\begin{equation}
\frac{1}{S}\mO\mY=
    \begin{bmatrix}
    p(0)   & \dots     & p(0) \\
    \vdots & \vdots & \vdots \\
    p(C)  & \dots & p(C) 
    \end{bmatrix}
    = \mP_{N}
    \label{eq:optim_sol}
\end{equation}
the substitution of (\ref{eq:optim_sol}) in (\ref{eq:simpl_2}) results in a function value of zero, which is the optimal distance value.
\end{IEEEproof}

The conditions described in Lemma \ref{lemm:opt_alloc} are quite restrictive. However, they reveal that distributing user groups equally can achieve low function values, which is the idea behind the heuristic in Algorithm  \ref{alg:heu}. If the allocated users are not equally distributed over the edges, the optimality condition will not be guaranteed. However, as long as our solution is close enough to the equal assignment, the proposed algorithm will still maintain a near-optimal performance, as will be shown in our evaluation. 
 
\section{Performance Evaluation}
\label{sec:eval}

In this section, we evaluate both, the \emph{optimization performance} and the \emph{learning performance} of the optimized assignment. In sub-section \ref{sec:evals1}, we evaluate the assignments configuration found by the different optimization approaches as measured by the resulting objective value. Then, in sub-section \ref{sec:evals2}, the assignment resulting from Algorithm \ref{alg:heu} is implemented and tested on several real-world data sets in a federated learning setting to measure the improvement in the learning speed and accuracy metrics.

\subsection{Optimization performance}
\label{sec:evals1}
To investigate the performance of the proposed optimization technique and the heuristic solution, we test these methods in a setup that contains $5$ edge servers and a varying number of groups whose sizes (i.e., number of users per group) are randomly generated in the range $0-10$. We test two scenarios: In the first one, the distribution over classes within each group is uniformly randomly generated ($0-10$ instances for each class), and these users are initially assigned to edge nodes randomly, resulting in an i.i.d data within each node. In the second case, The content of each group is generated such that the data within each group contains only $20\%$ of the total classes, and these users are initially assigned to edge nodes such that each edge contains only $20\%$ of the classes, making the data across edge nodes non-i.i.d.

\begin{figure}[t]
\centering
\includegraphics[width=0.485\textwidth]{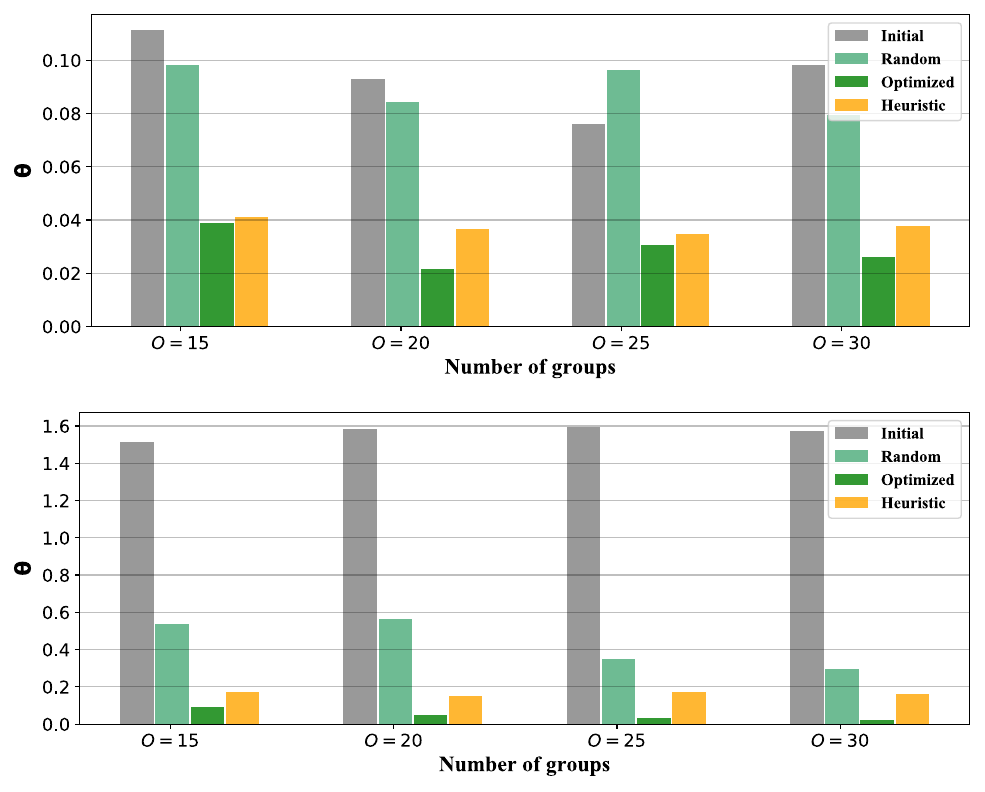}
\caption{Function values for different optimization techniques with initial i.i.d distribution (up), and non-i.i.d (down).}
\label{fig:opt}
\end{figure}

We use 10-class balanced datasets (i.e., $p(c=i)=0.1, \forall i$). Thus, in the non-i.i.d case, the users belonging to each node will contain $2$ classes only. The objective values are shown in Fig. \ref{fig:opt} for multiple assignment policies. The random policy allocates users to edge devices according to a uniform distribution. The optimized policy is the result of Branch and Bound solution to $\boldsymbol{P_2}$ after exploring $10^7$ nodes, and the heuristic policy is the assignment resulting from Algorithm \ref{alg:heu}. For this experiment only, we assume that users can be assigned to any of the edge nodes.

% can be assigned to all because of the feasibility and matlab implementation

In the i.i.d case, the initial value of the weighted average distribution distance (i.e., $\mathbf{\theta}$) is already close to the optimal. Furthermore, the random reassignment expectedly does not change the value much, as the data was non-i.i.d (shuffled) in the first place. Nonetheless, the optimized techniques still deliver better performance. In the designed non-i.i.d case, the initial $\theta$ is much higher, approximately $1.6$. This number comes from substituting the following values in $(\ref{eq:wavg_dist})$: $r^{(n)}=0.2$ (i.e., same data size for each edge), and $\|D^{(n)}\|_1=(0.5-0.1)+(0.5-0.1)+\times 8\times(0-0.1) = 1.6, \forall n$ (each node has instances of $2$ classes out of the $10$) . The number might not be exactly $1.6$ since the size of the proportion of data for each edge is not exactly $0.2$.

It can be seen that the random assignment already achieves a low total distribution distance. This is expected since users' uniform assignment will assign each edge with different users having samples from each class, resulting in an i.i.d data allocation. The proposed optimized approach still offers a lower value. This better performance is mainly because the randomization does not take into account the classes' distribution within each user and might allocate a user to any edge. However, the optimization search considers imbalances and allocates users in a way that is not necessarily uniform but guaranteed to have better probability distances (e.g., assigning users with an imbalanced and high number of samples of a specific class to edges with a deficit in that class). The proposed heuristic approach is an approximation of the optimization that delivers close performance. While it does not perform the expensive search, it greedily attempts equalization, aiming to provide each edge with all classes instances. This performs better than the random assignment (i.e., baseline). It is also identical to the optimal solution in cases where the classes are balanced, and the groups are divisible on the edges (as illustrated in Lemma 3). In the general case, however, such assignments obtain a performance close to the optimal solution.  
%% These assumption to access all are to be able to solve the optimization in few hours , should i say this? :). UPDATE: maybe i should because it is not tested in the main figure.

% Note, this number 10, could be anything. 
%30 is divisible by 2 and by 3. 

\begin{figure*}
\centering
\includegraphics[width=\textwidth]{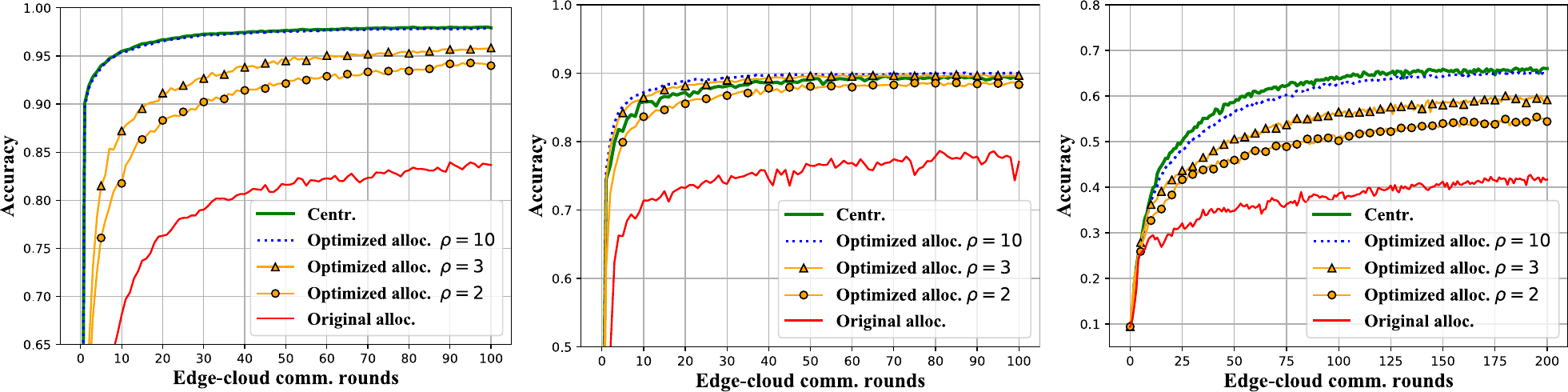}
\caption{Training performance per communication round for different NTCs using MNIST (left), FashionMNIST (middle), and CIFAR-10 (right) datasets.}
\label{fig:res}
\end{figure*}

\subsection{Learning performance}
\label{sec:evals2}
We provide comprehensive experiments to evaluate the learning performance under the proposed optimized assignments using three public datasets: MNIST \cite{mnist}, FashionMNIST \cite{fashion}, and CIFAR-10 \cite{cifar-10}. The neural architectures used for each dataset are provided in Appendix C. Although the neural networks' loss function does not satisfy the convexity assumption in lemma-2, they are considered for the empirical evaluation as the literature shows the benefits of balanced data even in non-convex optimization \cite{wang2019adaptive,zhao_federated_2018, Client-Edge-Cloud}. Herein, we consider $10$ edge nodes, each with $300$ users that we group into $10$ groups (each group with the same classes' distribution). The dataset is distributed such that each edge node only contains $2$ classes.

%Herein, we consider $10$ edge nodes, each with $10$ groups of size $30$. The dataset is distributed over users such that each node only contains $2$ classes.

Model accuracy vs. edge-cloud communication rounds is plotted in Fig. \ref{fig:res}; For each dataset, we plot $5$ models. The centralized model provides a benchmark for the performance since it is trained on the aggregated dataset of all users across all edges. The original assignment model performs the learning directly with the skewed user-edge assignment. The optimized assignment schemes perform the assignment according to Algorithm \ref{alg:heu}, assuming each user is in the proximity of $\rho$ edge nodes, before starting the learning process. The FL parameters are as follows: local epochs: $1$, local batch size: $10$, and learning rate $0.01$. For the centralized benchmark, the batch size is $100$. This is because the local batch size should be multiplied by the number of edge nodes so that at each communication round in the federated settings, the same number of data points are used for both the central and federated model \cite{zhao_federated_2018, mcmahan2017communication} 

\begin{table}[t]
\caption{Leaning performance indicators for different values of $\rho$}
\label{tab:stats}
\begin{tabular}{L|ccc|ccc}
\multicolumn{1}{l|}{} & \multicolumn{3}{c|}{Accuracy improvement} & \multicolumn{3}{c}{Speed improvement} \\ \cline{2-7} 
\multicolumn{1}{l|}{$\rho$} & MNIST & \begin{tabular}[c]{@{}c@{}}F.\\ MNIST\end{tabular} & CIFAR & MNIST & \begin{tabular}[c]{@{}c@{}}F.\\ MNIST\end{tabular} & CIFAR \\ \hline
$2$ & 12.53\% & 14.64\% & 30.53\% & 88\% & 96\% & 87\% \\ \hline
$3$ & 14.75\% & 16.42\% & 41.90\% & 93\% & 98\% & 89.5\% \\ \hline
$10$ & 17.03\% & 16.74\% & 56.06\% & 99\% & 98\% & 93\% \\ \hline
\end{tabular}
\end{table}

Across all datasets, the performance achieved by the optimized assignment when $\rho=10$ and the centralized benchmark are almost identical. This is because a full distribution of each group to all edge nodes is possible, resulting in an i.i.d data in each node. Hence, the initial weight divergence is expected to be close to zero, resulting in a similar performance to the centralized model, which keeps improving thereafter until convergence. Note that while $\rho=10$ is not necessarily practical, this experiment provides us with a validation to our expectation that when the distribution distance is close to $0$, a performance similar to the centralized gradient descent is expected. On the other hand, the original assignment model is slower to reach any specific accuracy and converges on a lower accuracy value. This is in line with what is found in the literature and is mainly due to the initial weight divergence and the overfitting of each edge to its local skewed distributions. The difference in the accuracy is most pronounced in the CIFAR dataset. CIFAR is known to be a more challenging classification problem compared to the other two, which explains the wider performance gap. Note that these accuracies do not represent the state-of-the-art. However, our focus in this paper is on the divergence between the federated models and their centralized version rather than achieving the most accurate centralized model. Hence, the neural architectures described earlier are sufficient for our purposes.

Table \ref{tab:stats} summarizes two important performance metrics:  The relative improvement in accuracy of the optimized assignments after model convergence compared to the initial assignment, and the communication round at which the optimized assignment reaches the same accuracy converged to by the original assignment. Regarding accuracy improvement, it can be seen that, in general, the performance improvements are the largest for CIFAR-10 dataset. Also, the majority of the improvement for all datasets is achieved for $\rho=2$. The difference in the performance gains across the dataset is due to the fact that we used the same FL parameters. Tuning neural architectures and other federated learning parameters per dataset is likely to affect the benefits of the optimized assignments. Regarding learning speed, all optimized models achieve non-i.i.d performance at much earlier communication rounds, demonstrating faster learning. For example, in CIFAR (Fig. \ref{fig:res}, right) the original allocation converges to an accuracy of $\sim41\%$ at episode $200$, while the optimized models reach the same (or first higher) value of accuracy at episodes $26$ for $\rho=2$, $21$ for $\rho=3$, and $14$ for $\rho=10$, yielding the improvements listed in the speed improvement column. The overall better performance metrics of the optimized assignment are due to each edge node having better representative samples of the global aggregated dataset, making their local gradients close, and hence their federated averaging more similar to the centralized one, as was discussed previously in the problem description.

\begin{figure}[t]
\centering
\includegraphics[width=0.48\textwidth]{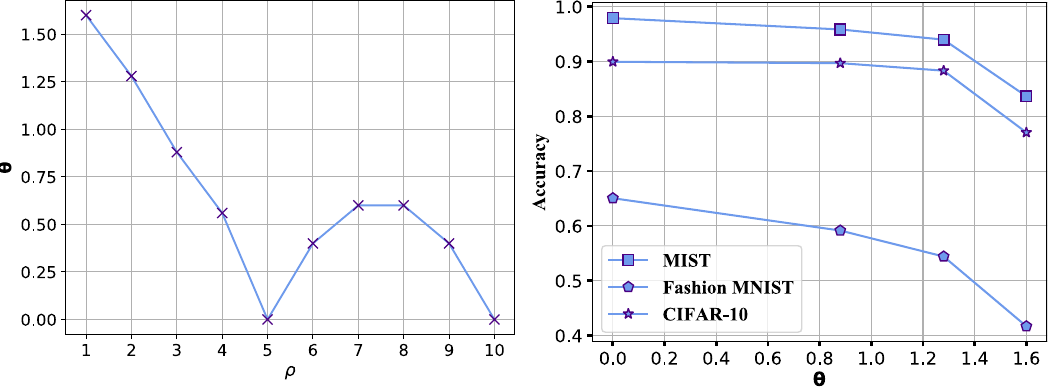}
\caption{The heuristic optimization performance for different values of $\rho$ (left), and the eventual accuracy obtained for different objective function values (right).}
\label{fig:dists_rhos}
\end{figure}

The proposed optimized assignment scheme is practical since it starts providing noticeable improvements with conservative values of $\rho$ (i.e., users need not be in the communication range of more than two edge nodes). The same trend of improved performance continues for the higher values of $\rho$, but with less performance gains over $\rho=2$. To study the effect of $\rho$, we plot the total distribution distance after executing Algorithm \ref{alg:heu} for different values of $\rho$ in Fig. \ref{fig:dists_rhos} (left).

When $\rho=1$ (i.e., no reassignment is possible), the value of the weighted distribution distance starts with $1.6$, as initially derived in the previous sub-section, then, increasing values of $\rho$ decreases the distribution distance for each node since nodes will see instances from new classes that it did not have before, reducing these classes' probability distance to the global population. The reduction continues until reaching $0$ for $\rho=5$, which occurs because an equal assignment of all groups, and hence class instances, for each node possible. The distance then increases again due to the imbalance resulting from the first $\rho$ nodes having more instances of some classes. Finally, a value of $\rho=10$ also satisfies the conditions presented in lemma-3, so an optimal value is feasible. Despite this irregular behavior for higher values of $\rho$, we are more concerned with the performance behavior for small values of $\rho$ since they represent a practical scenario (UEs are in proximity of $1$ or a small number of edges), for which consistent decrease in the total distribution distance, and hence performance improvement, can be expected.

%of satisfying the condition of lemma-3, making

We then plot the eventual accuracy obtained in our experiments on the three datasets for each value of $\boldsymbol{\theta}$ in Fig. \ref{fig:dists_rhos} (right). Both plots in Fig. \ref{fig:dists_rhos} explain the increased performance gains in the experiments with higher values of $\rho$; Higher values of $\rho$ enable lower objective function value (for small values of $\rho$), and the accuracy is highest for such lower objective values.
While the performance deterioration continues with higher values of $\boldsymbol{\theta}$, it does that at a faster rate. This is because neural networks are sensitive to initial weights and the early training stages (first few epochs) since they determine the trajectory of training after that \cite{goodfellow2016deep}.

%% describe the distance 
\subsection{Discussion/Limitations}
The proposed optimization focused on the edge-cloud communication round minimization as it is the main delay cause in HFL architecture. The edge-client network was assumed to provide a negligible communication delay. Thus, we did not consider the user-edge communication round. However, In some cases, the user-edge communication can contribute considerably to the delay, which requires federated learning, with extended local computation, also on the user-edge layer. In such a case, a conflicting dual-objective optimization arises: Assigning heterogeneous users to an edge will increase the user-edge distribution distance, but it will decrease the edge-cloud distribution distance since each edge is likely to contain all classes. Conversely, assigning similar users' to an edge will decrease the user-edge distribution distance, and increase the edge-cloud one. The solution to this dual-objective optimization is an entire Pareto front,  of which a certain point is considered optimal after determining the weight of each of the objectives (weight for the user-edge distribution distance minimization objective, and another for the edge cloud distribution distance minimization objective). This weight can be determined depending on multiple factors, such as the delay caused by each layer's topology and the energy constraints in each layer for user devices. The problem introduced in this paper can be considered as an instance of this dual-objective with no weight on the user-edge divergence. Interested readers may consider the concurrent works in \cite{yuan2020hierarchical}, which proposes P2P communication at the users' layer, and the convergence analysis in \cite{wang2020local} which studies the convergence rate of HFL considering FedAvg at both layers. In facts, the work in \cite{wang2020local} also suggests that careful ``grouping" strategies of UEs may enhance HFL performance, which is the subject of this paper.

While tackling multi-objective optimization in HFL is beyond the scope of this paper, it is a promising extension that generalizes the formulation and results found here.

\section{Conclusion}
\label{sec:conc}
In this paper, we studied the hierarchical federated learning paradigm, which leverages edge servers to synchronize users' models, and a cloud server to synchronize edges' models. The objective is to obtain accurate models with minimum edge-cloud communication rounds as they are the dominant source of delay in these systems. Toward this end, we first derived an expression for the upper bound of deviation of the federated model from the centralized one, which was shown to be proportional to the weighted distance between each node's distribution and the global one. Guided by this insight, we formalized an optimization problem that assigns users to edge nodes in their communication range to minimize the distribution distance between edge devices. However, the formulated problem is turned to be an NP-Hard problem. 

We then proposed two approaches to solve our problem. The first is a branch and bound-based solution to a simplified linear version of the problem that assumes equal assignments between edges. The second is a heuristic-based approach that greedily attempts to equalize class distributions among edge devices. Under certain conditions, we demonstrated that the two solution approaches yield the same optimal assignments. In the general case, both yield close sub-optimal solutions. We then tested the heuristic solution using three public datasets with results showing significant performance improvement when considering the users are in proximity of only two edge nodes. Eventually, we argue that both the communication properties of user devices and the statistical properties of their datasets should be considered in the design of future distributed and mobile learning systems.  To that end, promising future research can look at online algorithms to select participating clients based on their communication channels and quality of data as well as advanced aggregation algorithms (other than FedAvg) that are inherently robust to non-IID data (e.g., weighting based on factors other than the dataset size).
\section*{Acknowledgment}
This work was made possible by NPRP grant NPRP12S-0305-190231 from the Qatar National Research Fund (a member of Qatar Foundation). The findings achieved herein are solely the responsibility of the authors.

\bibliographystyle{IEEEtran}
\bibliography{references.bib}

\appendices

\section{lemma-1}
\label{app:centralized_federated_equality}
\begin{IEEEproof}
\begin {align}
\w^{(n)}_{t} &=\sum_{u\in \mathcal{U}_n} r^{(n,u)}\w^{(n,u)}_{t}\nonumber\\
\
&=\sum_{u\in \mathcal{U}_n} r^{(n,u)} \times \big(\w^{(n,u)}_{t-1} - \delta    \nabla \mathcal{L}^{(u)}(\w^{(n,u)}_{t-1})\big) \nonumber\\
\
&=\sum_{u\in \mathcal{U}_n} r^{(n,u)} \times \w^{(n,u)}_{t-1} - \sum_{u\in \mathcal{U}_n}  r^{(n,u)} \times \delta    \nabla \mathcal{L}^{(u)}(\w^{(n,u)}_{t-1}) \label{eq:edge_fed_weights}
\end {align}
 
The first term:
\begin{align}
    \sum_{u\in \mathcal{U}_n}  r^{(n,u)} \times \w^{(n,u)}_{t-1} &= \sum_{u\in \mathcal{U}_n} r^{(n,u)} \times \w^{(n)}_{t-1} &&\text{(due to  synch.)}\nonumber\\
    &=\w^{(n)}_{t-1} \label{eq:edge_loss_first_term}
\end{align}

The second term:
\begin{align}
     \sum_{u\in \mathcal{U}_n} r^{(n,u)}\times \delta    \nabla \mathcal{L}^{(u)}(\w^{(n,u)}_{t-1}) &= \delta   \nabla \sum_{u\in \mathcal{U}_n}  r^{(n,u)}
     \mathcal{L}^{(u)}(\w^{(n)}_{t-1}) \nonumber\\
     &=\delta    \nabla \mathcal{L}^{(n)}(\w^{(n)}_{t-1})\label{eq:edge_loss_linearity}
\end{align}
Where (\ref{eq:edge_loss_linearity}) holds because
\begin{align}
    \mathcal{L}^{(n)}(\w) &\doteq \frac{1}{|\cup_u \mathcal{D}_{n,u}|} \sum_{u\in \mathcal{U}_n} |\mathcal{D}_{n,u}| \mathcal{L}^{(u)} \nonumber\\
    &=\sum_{u\in \mathcal{U}_n} \frac{|\mathcal{D}_{n,u}|}{|\cup_u \mathcal{D}_{n,u}|} \mathcal{L}^{(u)} \nonumber \\
    &=\sum_{u\in \mathcal{U}_n} r^{(n,u)} \mathcal{L}^{(u)} \label{eq:edge_loss}
\end{align}

Substituting (\ref{eq:edge_loss_first_term}) and (\ref{eq:edge_loss_linearity}) in (\ref{eq:edge_fed_weights}), we obtain: 
\begin {align}
\w^{(n)}_t &= \underbrace{\w^{(n)}_{t-1} - \delta    \nabla \mathcal{L}^{(n)}(\w^{(n)}_{t-1})}_{\text{centralized GD using the loss on the virtual aggregated dataset}}
\end {align} \end{IEEEproof}

\section{Lemma-2}
\label{app:divergence_bound}
As mentioned earlier, the proof is based on steps similar to those in \cite{zhao_federated_2018}. However, we modify part-1 to accommodate HFL settings and add part-3 to write the divergence entirely in terms of $D^{(n)}$, which is necessary for our optimization formulation.

\begin{IEEEproof}
\subsection{Part 1}
In part $1$, we write $\lVert\w^{\text{f}}-\w\rVert$ in terms of $\lVert \w^{(n)}- \w \rVert$. 

\begin{align}
&\lVert\w^{\text{f}}_{mT}-\w_{mT}\rVert = \lVert \sum_{n=1}^Nr^{(n)}\w_{mT}^{(n)} -  \w_{mT} \rVert \nonumber \\
&=\lVert \sum_{n=1}^N  r^{(n)} \left( \w^{(n)}_{mT-1} - \delta \nabla_{\w^{(n)}}\mathcal{L}^{(n)}(\w^{(n)}_{mT-1}) \right)  \nonumber \\
&-\left( \w_{mT-1} - \delta \nabla_{\w}\mathcal{L}(\w_{mT-1}) \right) \rVert \nonumber \\
&=\lVert \sum_{n=1}^N  r^{(n)} \left( \w^{(n)}_{mT-1} - \delta \sum_{i=1}^C -p^{(n)}(i) J_i(\w^{(n)}_{mT-1})\right) \nonumber \\
&-\left( \w_{mT-1}- \delta \sum_{i=1}^C \sum_{n=1}^N - r^{(n)} p^{(n)}(i) J_i(\w_{mT-1}) \right) \rVert
\label{eq:losses_sub}
\end{align}
% below definition added in-body
%where $J_i(\w)=\nabla_{w} \mathbb{E}_{\boldsymbol{x}\sim q(\boldsymbol{x}|y=i)} \left[ \log d_i(\boldsymbol{x};\w) \right]$

(\ref{eq:losses_sub}) holds by substituting the loss functions, $\mathcal{L}(\cdot)$ and $\mathcal{L}^{(n)}(\cdot)$ by their definitions in (Eq. $1$ in the main paper) and (\ref{eq:edge_loss}), and using following equalities (from the probability multiplication rule):
\begin{equation}
p(i)=\sum_{n=1}^Nr^{(n)}p^{(n)}(i)
\end{equation}
and
\begin{equation}    
\sum_{u\in \mathcal{U}_n} r^{(n,u)} p^{(u)}(i)=p^{(n)}(i)
\end{equation}

By triangular inequality:
\begin{align}
\label{eq:weights_diff}
&(\ref{eq:losses_sub})\leq \lVert \sum_{n=1}^N  r^{(n)} \w^{(n)}_{mT-1}- \w_{mT-1} \rVert + \\
\label{eq:grads_diff}
& \delta\lVert \sum_{n=1}^N  r^{(n)}\sum_{i=1}^C p^{(n)}(i) \left(J_i(\w^{(n)}_{mT-1}) - J_i(\w_{mT-1})  \right) \rVert
\end{align}

The term in (\ref{eq:weights_diff})  $=\lVert \sum_{n=1}^N  r^{(n)} \w^{(n)}_{mT-1}-\sum_{n=1}^Nr^{(n)}\w_{mT-1} \rVert$ since it is a weighted sum of a $\w_{mT-1}$ with weights that sum to one  $\w_{mT-1}$. The gradient subtraction term is (\ref{eq:grads_diff}) is upper bounded by $L_i \times  \lVert \w^{(n)}_{mT-1} - \w_{mT-1} \rVert$ due to the $L_i-$Lipschitz assumption of $J_i$. The inequality becomes:
\begin{align}
\nonumber
&\lVert\w^{\text{f}}_{mT}-\w_{mT}\rVert \leq \sum_{n=1}^N  r^{(n)}  \lVert \w^{(n)}_{mT-1}- \w_{mT-1} \rVert + \\
&\delta\sum_{n=1}^N  r^{(n)}\sum_{i=1}^C p^{(n)}(i) L_i \times \lVert \w^{(n)}_{mT-1}- \w_{mT-1} \rVert
\end{align}

\begin{equation}
\label{eq:p1_result}
\lVert\w^{\text{f}}_{mT}-\w_{mT}\rVert \leq \sum_{n=1}^N  r^{(n)}  \lVert \w^{(n)}_{mT-1}- \w_{mT-1} \rVert \times P^{(n)}
\end{equation}
wehre $P^{(n)}=1 + \delta \sum_{i=1}^C p^{(n)}(i) \times L_i$

\subsection{Part 2}
In part 2, we derive an upper bound on $\lVert \w^{(n)}_{mT-1}- \w_{mT-1} \rVert$:

\begin{align}
\nonumber
&\lVert \w^{(n)}_{mT-1}- \w_{mT-1} \rVert \leq\\
\nonumber
&\lVert  \w^{(n)}_{mT-2} - \delta \sum_{i=1}^C p^{(n)}(i)  J_i(\w^{(n)}_{mT-2})  -\\
\nonumber
&\left( \w_{mT-2}- \delta \sum_{i=1}^C p(i) J_i(\w_{mT-2}) \right) \rVert \\
\nonumber
&\leq\lVert  \w^{(n)}_{mT-2} - \w_{mT-2} \rVert + \\
\nonumber
&\delta \lVert \sum_{i=1}^C p(i) J_i(\w_{mT-2}) - \sum_{i=1}^Cp^{(n)}(i)  J_i(\w^{(n)}_{mT-1}) \rVert
\end{align}
Regarding the second term of the inequality, we have that:
\begin{align}
\nonumber
&\delta \lVert \sum_{i=1}^C p(i) J_i(\w_{mT-2}) - \sum_{i=1}^C p^{(n)}(i)  J_i(\w^{(n)}_{mT-2}) \rVert\\
\nonumber
&\leq \delta \lVert \sum_{i=1}^C p(i) J_i(\w_{mT-2}) - \sum_{i=1}^C p^{(n)}(i)  J_i(\w^{(n)}_{mT-2}) \\
\nonumber
& +\sum_{i=1}^Cp^{(n)}(i)  J_i(\w_{mT-2}) - \sum_{i=1}^C p^{(n)}(i) J_i(\w_{mT-2}) \rVert\\
\nonumber
&\leq \delta \lVert \sum_{i=1}^C  \left( p(i) - p^{(n)}(i) \right) J_i(\w_{mT-2}) + \\
\nonumber
&\delta \sum_{i=1}^C p^{(n)}(i) \left( J_i(\w_{mT-2}) - J_i(\w^{(n)}_{mT-2})  \right)  \rVert
% the commin factor is X shape 
\\
\nonumber
&\leq \delta \lVert \sum_{i=1}^C  \left( p(i) -  p^{(n)}(i) \right) J_{max}(\w_{mT-2}) \rVert+\\
\nonumber
& \lVert\sum_{i=1}^C p^{(n)}(i) L (\w_{mT-1} - \w^{(n)}_{mT-1} )   \rVert
% the commin factor is X shape
\end{align}
% below definition added in-body
%where $J_{max}(\w) =\max_i J_i(\w) $.
The inequality (in part 2) becomes: 

%%%%%%%%%%%%%%%%%%%%%%%%%%%%%%%%%%%%%%%%%%%%%%%%%%%%%%%%%%%
\begin{align}
\nonumber
&\lVert \w^{(n)}_{mT-1}- \w_{mT-1} \rVert \leq \lVert  \w^{(n)}_{mT-2} - \w_{mT-2} \rVert&  \\
\nonumber
&+\delta \lVert \sum_{i=1}^C  \left( p(i) -  p^{(n)}(i)  \right) J_{max}(\w_{mT-2}) \rVert&\\
\nonumber
&+\lVert\sum_{i=1}^Cp^{(n)}(i) L (\w_{mT-1} - \w^{(n)}_{mT-1} )\rVert&
\end{align}
%the weight divergence common factor depends on the fact that their sign is irrelevent
\begin{align}
\nonumber
 \leq \left(1 + \delta \sum_{i=1}^C p^{(n)}(i) L_i \right)  \lVert  \w^{(n)}_{mT-2} - \w_{mT-2} \rVert + \\
\nonumber
\delta J_{max}(\w_{mT-2}) \underbrace{\sum_{i=1}^C \lVert \left( p(i) -  p^{(n)}(i) \right)}_{\|D^{(n)}\|_1}  \rVert
\end{align}
%%%%%%%%%%%%%%%%%%%%%%%%%%%%%%%%%%%%%%%%%%%%%%%%%%%%%%%%%%
\begin{multline}
\lVert \w^{(n)}_{mT-1}- \w_{mT-1} \rVert \leq P^{(n)} \times \lVert  \w^{(n)}_{mT-2} - \w_{mT-2} \rVert \\
+ \delta J_{max}(\w_{mT-2}) \times \|D^{(n)}\|_1)  \rVert
\end{multline}

We perform backward induction until the previous synchronization $(m-1)T$

\begin{align}
\nonumber
&\lVert \w^{(n)}_{mT-1}- \w_{mT-1} \rVert  \nonumber \\
& \leq P^{(n)} \times \lVert  \w^{(n)}_{mT-2} - \w_{mT-2} \rVert \nonumber \\ &+ \delta J_{max}(\w_{mT-2}) \times\|D^{(n)}\|_1)  \rVert \\ \nonumber
&\leq (P^{(n)})^2 \times \lVert  \w^{(n)}_{mT-3} - \w_{mT-3} \rVert  \nonumber \\
&+\delta  \left(  J_{max}(\w_{mT-3}) + P^{(n)}  J_{max}(\w_{mT-2}) \right) \times\|D^{(n)}\|_1 \nonumber\\
&\leq (P^{(n)})^{T-1} \times \lVert  \w^{(n)}_{(m-1)T} - \w_{(m-1)T} \rVert \nonumber \\
&+\delta  \sum_{h=0}^{T-2} \left(  (P^{(n)})^h   J_{max}(\w_{mT-h-2}) \right)
\times\|D^{(n)}\|_1 \label{eq:p2_result}
\end{align}

Noting that $w^{(n)}_{(m-1)T}=w^{(f)}_{(m-1)T}$ due to the synchronization, and  substituting the the result of part 2 (i.e., (\ref{eq:p2_result})) in the inequality of part 1 (i.e., (\ref{eq:p1_result})) yields:

\begin{multline}
\nonumber
\lVert\w^{\text{f}}_{mT}-\w_{mT}\rVert \leq \\
\sum_{n=1}^N  r^{(n)}  \bigg( (P^{(n)})^{T-1} \times  \lVert  \w^{\text{f}}_{(m-1)T} - \w_{(m-1)T} \rVert +\\
\delta  \sum_{h=0}^{T-2}   (P^{(n)})^h   J_{max}(\w_{mT-h-2})
\times\|D^{(n)}\|_1\bigg)  \times P^{(n)}
\end{multline}
Which yields the bound:

\begin{multline}
\label{eq:zhao_bound}
\lVert\w^{\text{f}}_{mT}-\w_{mT}\rVert \leq \sum_{n=1}^N  r^{(n)} P^{T}   \lVert  \w^{\text{f}}_{(m-1)T} - \w_{(m-1)T} \rVert \\
+\delta  \sum_{n=1}^N  r^{(n)} \sum_{h=0}^{T-2} \left(  (P^{(n)})^{h+1}   J_{max}(\w_{mT-h-2})\right)\times\|D^{(n)}\|_1
\end{multline}

\subsection{Part 3}
In part 3, we write the upper bound of $\lVert\w^{\text{f}}_{mT}-\w_{mT}\rVert$ in terms of $\|D^{(n)}\|_1$.
Note that the bound derived in (\ref{eq:zhao_bound}) has two terms, a recursive term, and a term that depends on $\|D^{(n)}\|_1$. In this part, we use back induction to write the bound as one term depending on $\|D^{(n)}\|_1$, which will give insights about their relationship.

By substituting $\lVert  \w^{\text{f}}_{(m-1)T} - \w_{(m-1)T} \rVert$ by its upper bound, (\ref{eq:zhao_bound}) can be written as:
\begin{align}
&\lVert\w^{\text{f}}_{mT}-\w_{mT}\rVert \nonumber \\
&\leq \sum_{n=1}^N  r^{(n)} (P^{(n)})^{T}  \times \big \lVert \sum_{n=1}^N  r^{(n)} (P^{(n)})^{T}    \lVert \w^{\text{f}}_{(m-2)T} - \w_{(m-2)T} \rVert \nonumber \\
&+\delta  \sum_{n=1}^N  r^{(n)} \sum_{h=0}^{T-2}   (P^{(n)})^{h+1}   J_{max}(\w_{(m-1)T-h-2}) \times \|D^{(n)}\|_1  \big\rVert \nonumber \\
&+ \delta  \sum_{n=1}^N  r^{(n)} \sum_{h=0}^{T-2}   (P^{(n)})^{h+1}   J_{max}(\w_{mT-h-2}) \times \|D^{(n)}\|_1 \nonumber
\end{align}

By similar backward induction until the initial weights (i.e., $m=0$), and assuming that the weights were equally initialized (i.e., $\w^{\text{f}}_0 = \w_0$ )

\begin{multline}
\lVert\w^{\text{f}}_{mT}-\w_{mT}\rVert \leq
\sum_{v=0}^{m}  \left( \sum_{n=1}^N  r^{(n)} (P^{(n)})^{T} \right)^v \nonumber \\
\times \delta  \sum_{n=1}^N  r^{(n)} \sum_{h=0}^{T-2}   (P^{(n)})^{h+1}   J_{max}(\w_{(m-v)T-h-2}) \times \|D^{(n)}\|_1
\end{multline}
\end{IEEEproof}
\section{}
% update below, add relu to fmnist in code
Neural networks architecture used for each dataset are shown below. ReLU activation is used after fully connected and pooling layers.
\subsection{MNIST}
Fully connected $(784\times200)$ $\to$ Dropout $(p=0.5)$ $\to$ Fully connected $(200,10)$ $\to$ Softmax

\subsection{F-MNIST}
(Using the convention $(in\_channels,out\_channels)$, kernel) For convolutional layers.)

The CNN model is: Convolutional: $(1,8)$, kernel $(3,3)$ $\to$ MaxPool $(3,3)$  $\to$ Convolutional $8, 16$ kernel $(2,2)$ $\to$  MaxPool $(3,3)$  $\to$ Fully connected $(576\times120)$ $\to$ Dropout $(p=0.25)$ $\to$ Fully connected $(120,60)$ $\to$ Fully connected $(60,10)$ $\to$ Softmax

\subsection{CIFAR}
The CNN model is: Convolutional: $(3,6)$, kernel $(5,5)$ $\to$ MaxPool $(2,2)$  $\to$ Convolutional $(6, 16)$ kernel $(5,5)$  $\to$ MaxPool $(2,2)$ $\to$ Fully connected $(400\times120)$ $\to$ Dropout $(p=0.5)$ $\to$ Fully connected $(120,84)$ $\to$ Dropout $(p=0.5)$ $\to$ Fully connected $(84,10)$ $\to$ Softmax

(Similar to the architecture in the PyTorch tutorial \cite{Pytorch_training}
\end{document}